\documentclass{article}
\usepackage{spconf,amsmath,graphicx}

\usepackage{cleveref} 
\usepackage[table]{xcolor} 
\usepackage{pifont} 
\newcommand{\cmark}{\ding{51}} 

\usepackage{amsfonts}
\usepackage{url}

\title{WHEN SIMPLICITY WINS: BOTTLENECK-AWARE CONTEXT MODELING FOR LIGHTWEIGHT SEMANTIC SEGMENTATION}
\name{Mian Muhammad Naeem Abid, Nancy Mehta, Zongwei Wu, Radu Timofte\thanks{This work was partially supported by the Humboldt Foundation.}}
\address{Computer Vision Lab, CAIDAS \& IFI, University of Würzburg, Germany}

\usepackage{etoolbox}


\usepackage{wrapfig}
\usepackage{multirow}
\usepackage{enumitem}

\begin{document}

\maketitle

\begin{abstract}
Semantic segmentation demands a careful balance between accuracy, efficiency, and scalability, which remains difficult to achieve for high-resolution imagery. Convolutional networks effectively model local patterns but struggle with long-range dependencies, whereas Vision Transformers capture global context at a high computational cost. While recent work largely focuses on encoder design, the bottleneck stage—central to contextual aggregation and information flow—has been relatively overlooked. We propose SiConMo, a lightweight yet effective framework, implemented in two variants: an RGB-only model (SiConMo) and a GME-enhanced variant (SiConMo$_\dagger$). We show that simplicity arises from a key design principle: at very low computational budgets, the bottleneck is the most efficient stage to integrate local and global context. SiConMo integrates three complementary components: a Token Pyramid Extraction Module for hierarchical multi-scale representation, a Transformer-Branched Depthwise Convolution block for bottleneck-aware context modeling, and a Feature Merging Module that preserves spatial structure while enhancing semantic consistency. Extensive experiments on ADE20K, PASCAL Context, Cityscapes, and COCO-Stuff demonstrate that SiConMo achieves a state-of-the-art accuracy--efficiency trade-off among lightweight semantic segmentation models, highlighting simplicity as a powerful design principle.
\url{https://github.com/miannaeem-lab/SiConMo}
\end{abstract}

\begin{keywords}
Lightweight Segmentation, Vision Transformers, Convolutional Neural Networks, Simplicity, Context Modeling, Efficiency
\end{keywords}

\section{Introduction}
\label{sec:intro}
\begin{figure}[t]
  \centering
\includegraphics[width=.8\linewidth]{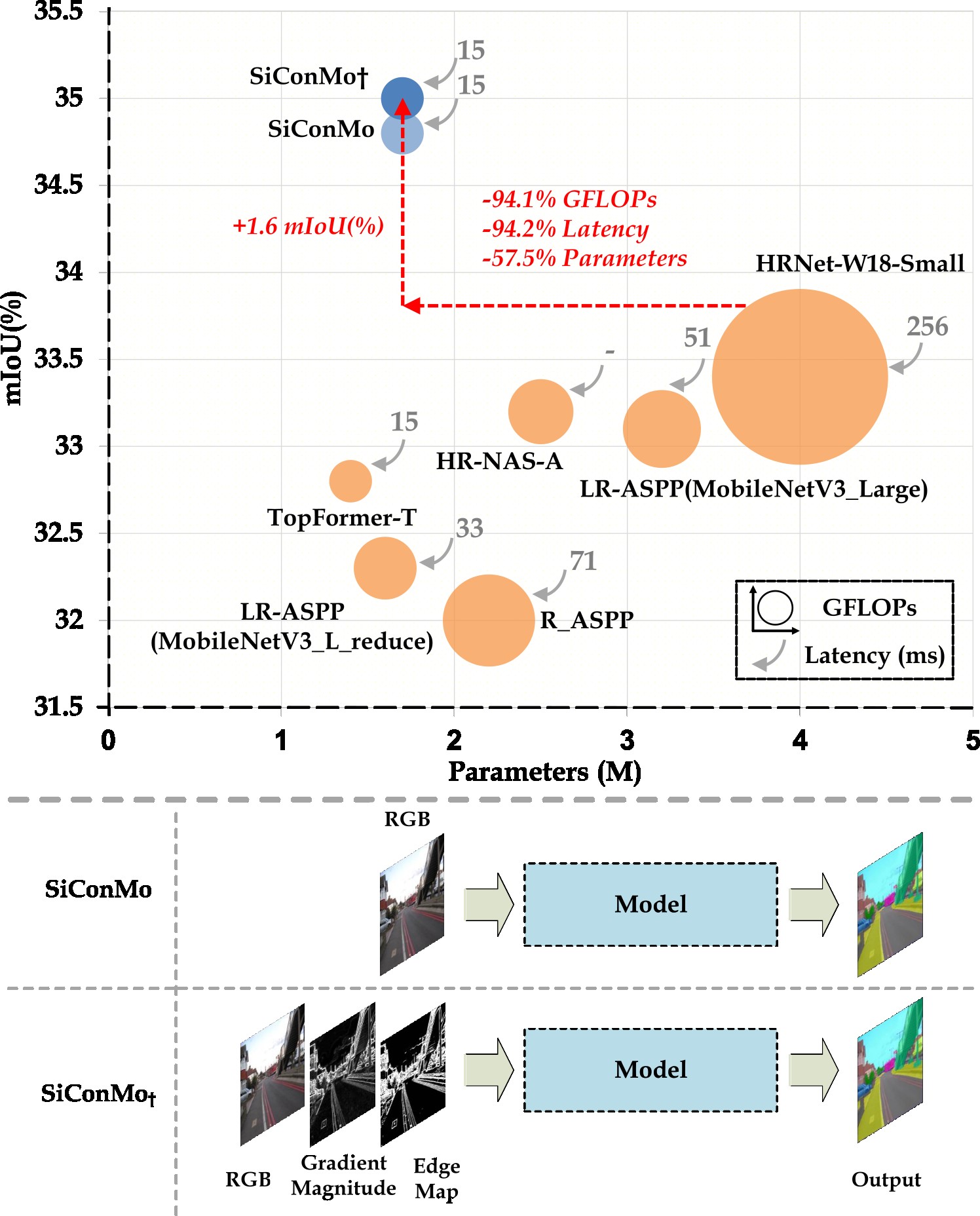}
 \caption{\textit{(Top)} Efficiency–accuracy comparison of lightweight semantic segmentation models on ADE20K validation set. Circle sizes denote GFLOPs. SiConMo achieves a superior trade-off, illustrating the effectiveness of a simple bottleneck-aware design principle. \textit{(Bottom)} Variants of the proposed SiConMo architecture.}
  \label{fig:comparison_of_all_models}
  \vspace{-3mm}
\end{figure}

Semantic segmentation is a fundamental computer vision task that requires dense, pixel-level prediction in complex visual scenes~\cite{mo2022review}. Convolutional Neural Networks (CNNs) have historically dominated this area due to their strong local feature modeling capabilities, but their limited receptive fields restrict long-range dependency modeling. Vision Transformers (ViTs) address this limitation through global self-attention~\cite{xie2021segformer}, yet their quadratic computational complexity makes them impractical for high-resolution and real-time segmentation.

To reduce the overhead of ViTs, prior work explores localized or window-based attention, separable attention mechanisms, and aggressive spatial downsampling~\cite{li2022efficientformer, pan2022edgevits}. While computationally efficient, these strategies often degrade segmentation accuracy by limiting receptive fields or discarding fine-grained spatial information. Conversely, lightweight CNNs based on depthwise separable convolutions~\cite{sandler2018mobilenetv2} scale efficiently but struggle to capture global context, resulting in suboptimal performance on complex datasets.

Recent hybrid CNN–Transformer models attempt to balance efficiency and expressiveness~\cite{zhang2022topformer}, yet most designs emphasize encoder complexity while overlooking the bottleneck stage, where contextual aggregation and feature redistribution naturally occur at minimal computational cost. Transformer-based bottlenecks often introduce redundant global attention, whereas CNN-based bottlenecks remain restricted to local receptive fields. This suggests that the bottleneck is a uniquely efficient point for mixing global and local context, but it remains underutilized in existing designs.

This work challenges the assumption that increasing encoder complexity is the primary path to improved performance.
Instead, we show that simplicity, when placed at the right architectural location, can be more effective than over-engineering.
We propose SiConMo, a lightweight bottleneck-aware framework built on this design principle. We further evaluate an enhanced variant, SiConMo$_\dagger$, which incorporates additional Gradient Magnitude and Edge Maps (GME) to improve structural awareness.
SiConMo integrates local and global feature processing without unnecessary architectural depth. It consists of (1) a Token Pyramid Extraction Module (TPEM) for efficient multi-scale token representation, (2) a Transformer-Branched Depthwise Convolution (Trans-BDC) block that enables lightweight context mixing at the bottleneck, and (3) a Feature Merging Module (FMM) that preserves spatial structure while enhancing semantic consistency.
Together, these components form a lightweight architecture that achieves a favorable efficiency–accuracy trade-off, as shown in \Cref{fig:comparison_of_all_models}.

We conduct a comprehensive and rigorous evaluation of SiConMo across multiple challenging benchmarks.
Our results consistently show that SiConMo achieves a strong and reliable balance between segmentation accuracy and computational efficiency, making it well-suited for real-time and resource-constrained applications.

\noindent Our contributions are as follows:
\begin{itemize}[itemsep=1pt, topsep=1pt, leftmargin=*]
\item We introduce SiConMo, a lightweight bottleneck-aware framework that demonstrates how simplicity in design, when applied at the bottleneck stage, can rival more complex architectures.
\item We propose and integrate the Transformer-Branched Depthwise Convolution (Trans-BDC) block, together with a Token Pyramid Extraction Module and a Feature Merging Module, enabling efficient multi-scale representation and lightweight context mixing at minimal computational cost.
\item We extensively evaluate SiConMo on challenging segmentation benchmarks, including ADE20K~\cite{zhou2017scene}, PASCAL Context~\cite{mottaghi2014role}, Cityscapes~\cite{cordts2016cityscapes}, and COCO-Stuff~\cite{caesar2018coco}, and further demonstrate its generalization capability on COCO object detection under real-time efficiency constraints.
\end{itemize}
\vspace{-3mm}

\section{Related Work}
\label{sec:related_work}

\noindent \textbf{Efficient Convolutional Neural Networks:}
Convolutional Neural Networks (CNNs) have long been central to visual recognition due to their strong inductive biases~\cite{he2016deep}. However, their computational cost limits deployment in resource-constrained settings. Lightweight CNNs such as MobileNet~\cite{howard2017mobilenets, sandler2018mobilenetv2}, ShuffleNet~\cite{ma2018shufflenet}, and GhostNet~\cite{han2020ghostnet} address this issue through depthwise separable convolutions and efficient designs. For semantic segmentation, approaches like DFANet~\cite{li2019dfanet} further reduce complexity via multi-scale feature aggregation. These methods, however, primarily emphasize local context modeling. In contrast, SiConMo enhances context representation by integrating multi-scale token processing and adaptive feature redistribution within a lightweight bottleneck.

\noindent \textbf{Lightweight Vision Transformers:}
Vision Transformers (ViTs) enable global context modeling through self-attention but incur high computational overhead. To improve efficiency, models such as LeViT~\cite{graham2021levit} and EfficientFormer~\cite{li2022efficientformer} introduce architectural and training optimizations or combine convolutional and transformer components. While these designs balance efficiency and accuracy, many rely on complex attention mechanisms. In contrast, SiConMo focuses on a simple, bottleneck-aware design that captures both local and global context with minimal computational overhead.
\vspace{-3mm}

\section{Proposed Method}
\label{sec:proposed_method}

\begin{figure*}[t]
  \centering
  \includegraphics[width=\linewidth]{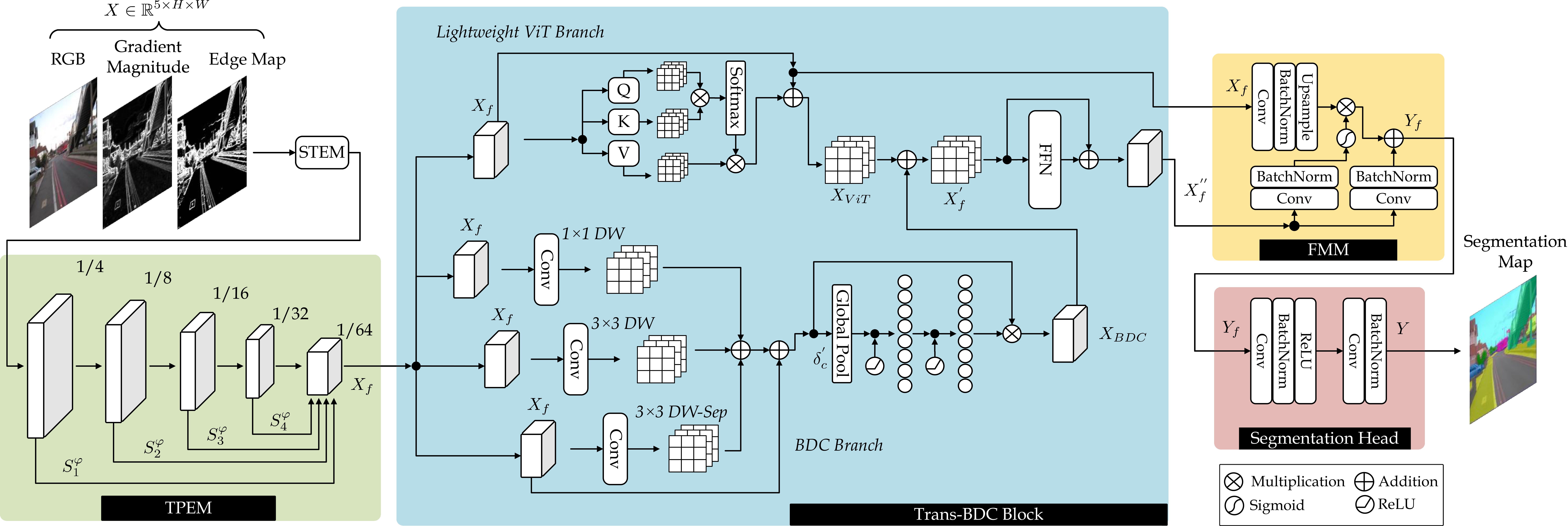}
  \caption{Overview of the proposed SiConMo framework. The architecture integrates the Token Pyramid Extraction Module (TPEM) for multi-scale representation, the Trans-BDC block for bottleneck-aware context modeling, and the Feature Merging Module (FMM) for adaptive spatial–semantic integration.}
  \label{fig:proposed_method}
  \vspace{-4mm}
\end{figure*}

We propose SiConMo, a simple yet effective, lightweight, bottleneck-aware framework for semantic segmentation that balances representational richness with computational efficiency. Rather than relying on heavy backbones or complex multi-stage architectures, SiConMo demonstrates that carefully designed lightweight modules can effectively capture both local structural details and global contextual dependencies. As illustrated in \Cref{fig:proposed_method}, the framework consists of three key components: (1) the Token Pyramid Extraction Module (TPEM), (2) the Trans-BDC bottleneck block, and (3) the Feature Merging Module (FMM).

\subsection{Token Pyramid Extraction Module}
The Token Pyramid Extraction Module (TPEM) constructs hierarchical multi-scale representations from the input image. Depending on the variant, the input is either RGB alone or RGB concatenated with gradient magnitude and edge maps to enrich structural cues. TPEM employs a sequence of inverted residual blocks from MobileNetV2~\cite{sandler2018mobilenetv2, zhang2022topformer} to generate feature maps $\{S_1, S_2, S_3, S_4\}$ at progressively coarser resolutions, where each feature map has size $\frac{H}{2^{i+1}} \times \frac{W}{2^{i+1}} \times C_i$ for $i \in \{1,2,3,4\}$.

To reduce computational cost while preserving contextual diversity, each feature map is pooled and concatenated along the channel dimension:
\begin{equation}
X_f = \langle S_1^{\varphi}, S_2^{\varphi}, S_3^{\varphi}, S_4^{\varphi} \rangle ,
\end{equation}
where $\varphi$ denotes average pooling. This operation efficiently combines fine-grained local details with coarse global context, producing a compact representation suitable for bottleneck processing.

\subsection{Trans-BDC Block: The Novel Bottleneck} 
The Trans-BDC block forms the core bottleneck of SiConMo, integrating local and global feature modeling within a single lightweight module. Unlike conventional bottlenecks that rely solely on convolutions or self-attention, Trans-BDC combines both paradigms through two parallel branches.

\noindent\textbf{Branched Depthwise Convolution (BDC) Branch:}
This branch captures localized spatial patterns using three parallel operations: a $3\times3$ depthwise convolution ($\xi_{3\times3}^{dw}$) for neighborhood modeling, a $1\times1$ depthwise convolution ($\xi_{1\times1}^{dw}$) for lightweight per-channel refinement, and a depthwise separable convolution ($\xi_{1\times1}^{pw}(\xi_{3\times3}^{dw}(.)$) for efficient feature fusion. The aggregated output is:
\begin{equation}
\delta_c' = \xi_{3\times3}^{dw}(X_f) + \xi_{1\times1}^{dw}(X_f) + \xi_{1\times1}^{pw}(\xi_{3\times3}^{dw}(X_f)) + X_f ,
\end{equation}
which is further refined via channel attention:
\begin{equation}
X_{BDC} = \Gamma_2(\Gamma_1(\varphi_g(\delta_c'))) \, \bar{\otimes} \, \delta_c' .
\end{equation}
Here, $\varphi_g$ denotes global average pooling, $\Gamma_1$ and $\Gamma_2$ are fully connected layers, and $\bar{\otimes}$ represents the Hadamard product.

\noindent\textbf{Lightweight ViT Branch:}
In parallel, a lightweight transformer branch models long-range dependencies using self-attention applied to pooled tokens. Efficiency is achieved through low-dimensional projections for queries, keys, and values, $1\times1$ convolutions in place of standard MLPs, batch normalization, and ReLU6 activations:
\begin{equation}
\begin{split}
X_{ViT} &= Attention(X_f) + X_f, \\
Attention &= \text{softmax}\left(\frac{QK^T}{\sqrt{d_k}}\right)V .
\end{split}
\end{equation}

The outputs of both branches are fused and passed through a depthwise-enhanced Feed-Forward network (FFN):
\begin{equation}
X_f' = X_{ViT} + X_{BDC}, \quad
X_f'' = FFN(X_f') + X_f' .
\end{equation}
This design shows that a carefully constructed hybrid bottleneck can replace more complex architectures while maintaining strong segmentation performance.

\vspace{-2.5mm}
\subsection{Feature Merging Module and Segmentation Head}
The Feature Merging Module (FMM) integrates local features from TPEM with global features from the Trans-BDC block using a gated fusion mechanism~\cite{zhang2022topformer}. Local features are projected via a $1\times1$ convolution ($\xi_{1\times1}^{c1}$), while global features are adaptively weighted through a sigmoid-activated $1\times1$ convolution, $\sigma(\xi_{1\times1}^{c2}(.))$. The merged feature map is:
\begin{equation}
Y_f = (\xi_{1\times1}^{c1}(X_f)) \, \bar{\otimes} \, (\sigma(\xi_{1\times1}^{c2}(X_f''))) + \xi_{1\times1}^{c3}(X_f'') .
\end{equation}
The fused representation is upsampled to the target resolution and processed by a lightweight segmentation head consisting of two $1\times1$ convolutions to produce the final prediction $Y$.

\vspace{-3.5mm}
\subsection{Variants and Design Philosophy}
SiConMo is implemented in two variants: an RGB-only configuration and an enhanced version incorporating Gradient Magnitude and Edge Maps (GME), denoted as SiConMo and SiConMo$_\dagger$.
Across all components, SiConMo follows a minimalistic design philosophy: preserving multi-scale representations without deep encoders, employing lightweight hybrid bottlenecks instead of full transformers, and using simple yet effective feature fusion. These choices highlight that careful bottleneck-aware design can achieve strong accuracy and efficiency without architectural over-complexity.
\vspace{-2mm}

\section{Experiments}
\label{sec:experiments}

\begin{table*}[t]
\small
\caption{Quantitative evaluation on ADE20K~\cite{zhou2017scene} and complementary analyses.
(a) Comparison with state-of-the-art representative semantic segmentation models on ADE20K dataset.
(b) Input-resolution variants trained at $448\times448$.
(c) Ablation study on ADE20K.
(d) GME variants on ADE20K.
GFLOPs are reported at $512\times512$ input resolution with single-scale inference; `-' denotes unreported values.
(e) object detection results on COCO dataset using RetinaNet (Sec.~\ref{subsec:object_detection}).
SiConMo$_\dagger$: GME variant.}
\label{tab:ade20k_mega}
\centering
\setlength{\tabcolsep}{2.5pt}
\renewcommand{\arraystretch}{0.85}
\begin{minipage}[t]{0.60\textwidth}
\vspace{0pt}
\centering
{\bf (a) Results on ADE20K}\par\vspace{0.5mm}
\vtop{\hbox{%
\small
\resizebox{\textwidth}{!}{%
\begin{tabular}{c|c|c|c|c|c}
\hline
Models & Encoder & mIoU & GFLOPs & Params & Latency(ms)\\
\hline
\multicolumn{6}{c}{\textbf{Heavyweight Models}}\\
\hline
PSPNet~\cite{zhao2017pyramid} & MobileNetV2~\cite{sandler2018mobilenetv2} & 29.6 & 52.2 & 13.7M & 426 \\
FCN-8s~\cite{long2015fully} & MobileNetV2~\cite{sandler2018mobilenetv2} & 19.7 & 39.6 & 9.8M & 406 \\
Semantic FPN~\cite{kirillov2019panoptic} & ConvMLP-S~\cite{li2023convmlp} & 35.8 & 33.8 & 12.8M & 311 \\
DeepLabV3+~\cite{chen2018encoder} & EfficientNet~\cite{tan2019efficientnet} & 36.2 & 26.9 & 17.1M & 388 \\
DeepLabV3+~\cite{chen2018encoder} & MobileNetV2~\cite{sandler2018mobilenetv2} & 38.1 & 25.8 & 15.4M & 414 \\
Lite-ASPP~\cite{chen2018encoder} & ResNet18~\cite{he2016deep} & 37.5 & 19.2 & 12.5M & 259 \\
PEM~\cite{cavagnero2024pem} & STDC1~\cite{cavagnero2024pem} & 39.6 & 16.0 & 17.0M & - \\
DeepLabV3+~\cite{chen2018encoder} & ShuffleNetv2-1.5x~\cite{ma2018shufflenet} & 37.6 & 15.3 & 16.9M & 384 \\
HRNet-Small~\cite{yuan2020object} & HRNet-W18-Small~\cite{yuan2020object} & 33.4 & 10.2 & 4.0M & 256 \\
SegFormer~\cite{xie2021segformer} & MiT-B0~\cite{xie2021segformer} & 37.4 & 8.4 & 3.8M & 308 \\
FeedFormer-B0~\cite{shim2023feedformer} & MiT-B0~\cite{xie2021segformer} & 39.2 & 7.8 & 4.5M & - \\
SegNeXt~\cite{guo2022segnext} & SegNeXt-T~\cite{guo2022segnext} & 41.1 & 6.6 & 4.3M & - \\
U-MixFormer~\cite{yeom2025u} & MiT-B0~\cite{xie2021segformer} & 41.2 & 6.1 & 6.1M & - \\
\hline
\multicolumn{6}{c}{\textbf{Lightweight Models}}\\
\hline
Lite-ASPP~\cite{chen2018encoder} & MobileNetV2~\cite{sandler2018mobilenetv2} & 36.6 & 4.4 & 2.9M & 94 \\
R-ASPP~\cite{sandler2018mobilenetv2} & MobileNetV2~\cite{sandler2018mobilenetv2} & 32.0 & 2.8 & 2.2M & 71 \\
HR-NAS-B~\cite{ding2021hr} & Searched~\cite{ding2021hr} & 34.9 & 2.2 & 3.9M & - \\
LR-ASPP~\cite{howard2019searching} & MobileNetV3-Large~\cite{howard2019searching} & 33.1 & 2.0 & 3.2M & 51 \\
HR-NAS-A~\cite{ding2021hr} & Searched~\cite{ding2021hr} & 33.2 & 1.4 & 2.5M & - \\
LR-ASPP~\cite{howard2019searching} & MobileNetV3-Large-reduce~\cite{howard2019searching} & 32.3 & 1.3 & 1.6M & 33 \\
LeMoRe~\cite{11084523} & LeMoRe~\cite{11084523} & 32.7 & 0.8 & 1.6M & 24 \\
TopFormer~\cite{zhang2022topformer} & TopFormer-T~\cite{zhang2022topformer} & 32.8 & 0.6 & 1.4M & 16 \\
SeaFormer~\cite{wan2023seaformer} & SeaFormer-T~\cite{wan2023seaformer} & 34.6 & 0.6 & 1.7M & 16 \\
\hline
\rowcolor{gray!20}
SiConMo (Ours) & Ours & 34.8 & 0.6 & 1.7M & 15 \\
\rowcolor{gray!20}
SiConMo$_\dagger$ (Ours) & Ours & 35.0 & 0.6 & 1.7M & 15 \\
\hline
\end{tabular}
}}%
}

\vspace{1.2mm}
{\bf (f) Object detection performance of SiConMo$_\dagger$ on COCO evaluation metrics}\par\vspace{0.5mm}
\vtop{\hbox{%
\makebox[\linewidth][c]{%
\small
\begin{tabular}{c|c|c|c|c|c|c}
\hline
Backbone & mAP & AP50 & AP75 & APS & APM & APL \\
\hline
\rowcolor{gray!20}
SiConMo$_\dagger$ (Ours)& 31.6 & 51.1 & 32.8 & 17.5 & 33.9 & 43.0 \\
\hline
\end{tabular}
}%
}}

\end{minipage}\hfill
\begin{minipage}[t]{0.39\textwidth}
\vspace{0pt}
\centering
{\bf (b) SiConMo variants trained with $448^2$ input size}\par\vspace{0.5mm}
\vtop{\hbox{%

\small
\resizebox{\textwidth}{!}{%
\begin{tabular}{c|c|c|c|c|c}
  \hline
  Models & Input & mIoU & GFLOPs & Params & Latency(ms)\\
  \hline
  \rowcolor{gray!20}
  SiConMo* & $448 \times 448$ & 34.4 & 0.5 & 1.7M & 14\\
  \rowcolor{gray!20}
  SiConMo$_\dagger$* & $448 \times 448$ & 34.7 & 0.5 & 1.7M & 14\\
  \hline
\end{tabular}
}}%
}

\vspace{1.2mm}
{\bf (c) Ablation Study of SiConMo on ADE20K}\par\vspace{0.5mm}
\vtop{\hbox{%
\small
\resizebox{\textwidth}{!}{%
\begin{tabular}{c|c|c|c|c|c|c|c|c}
  \hline
  \multicolumn{6}{c|}{SiConMo (Trans-BDC)} & mIoU & GFLOPs & Params\\
  \cline{1-6}
  ViT$_{block}$ & $3\times3_{dw}$& $1\times1_{dw}$& $3\times3_{dw.sep}$& C-Attn & GME &  &  &  \\
  \hline
    & \cmark &  &  &  &  & 25.4 & 0.55 & 1.02M\\
    & \cmark & \cmark &  &  &  & 25.6 & 0.55 & 1.10M\\
    & \cmark & \cmark & \cmark &  &  & 28.7 & 0.56 & 1.29M\\
    & \cmark & \cmark & \cmark & \cmark &  & 29.5 & 0.56 & 1.38M\\
    & \cmark & \cmark & \cmark & \cmark & \cmark & 29.8 & 0.56 & 1.38M\\
  \hline
  \cmark &  &  &  &  &  & 32.7 & 0.54 & 1.41M\\
  \cmark & \cmark &  &  &  &  & 32.8 & 0.57 & 1.42M\\
  \cmark & \cmark & \cmark &  &  &  & 32.9 & 0.57 & 1.43M\\
  \cmark & \cmark & \cmark & \cmark &  &  & 33.8 & 0.58 & 1.61M\\
  \rowcolor{gray!20}
  \cmark & \cmark & \cmark & \cmark & \cmark &  & 34.8 & 0.58 & 1.68M\\
  \rowcolor{gray!20}
  \cmark & \cmark & \cmark & \cmark & \cmark & \cmark & 35.0 & 0.58 & 1.68M\\
  \hline
\end{tabular}
}}%
}

\vspace{1.2mm}
{\bf (d) Comparison of SiConMo GME variants}\par\vspace{0.5mm}
\vtop{\hbox{%
\small
\resizebox{\textwidth}{!}{%
\begin{tabular}{c|c|c|c|c}
  \hline
  GME (Variations) & mIoU & GFLOPs & Params & Latency(ms)\\
  \hline
  Canny & 31.6 & 0.6 & 1.7M & 18.8\\
  Sobel+Canny & 35.0 & 0.6 & 1.7M & 19.2\\
  \rowcolor{gray!20}
  Sobel (Ours) & 35.0 & 0.6 & 1.7M & 15.1\\
  \hline
\end{tabular}
}}%
}

\vspace{1.2mm}
{\bf (e) Object detection results on COCO dataset}\par\vspace{0.5mm}
\vtop{\hbox{%
\makebox[\linewidth][c]{%
\small
\begin{tabular}{c|c|c|c}
  \hline
Backbone & mAP & GFLOPs & Parameters  \\
 \hline
ShuffleNetV2 \cite{ma2018shufflenet} & 25.9 & 161 & 10.4M \\
MobileNetV3 \cite{howard2019searching} & 27.2 & 162 & 12.3M \\
TopFormer-T \cite{zhang2022topformer} & 27.1 & 160 & 10.5M \\
SeaFormer-T \cite{wan2023seaformer} & 31.5 & 160 & 10.9M \\
\hline
\rowcolor{gray!20}
SiConMo$_\dagger$ (Ours) & 31.6 & 160 & 10.9M \\
\hline
\end{tabular}
}%
}}

\end{minipage}
\vspace{-5mm}
\end{table*}
\vspace{-2mm}
\subsection{Results on ADE20K}
\label{sec:results_ade20k}
\vspace{-2mm}
\Cref{tab:ade20k_mega}(a) reports a quantitative comparison of SiConMo against representative state-of-the-art semantic segmentation methods on the ADE20K validation set.
Compared with the higher-capacity transformer-based U-MixFormer model using a MiT-B0 encoder, SiConMo achieves competitive segmentation accuracy while reducing computational cost by 90.2\% and parameters by 72.1\%.
This efficiency is primarily enabled by its bottleneck-aware design, which combines multi-scale token representations with lightweight hybrid context modeling instead of full self-attention.
Compared with CNN-based lightweight approaches such as LR-ASPP with MobileNetV3-Large, SiConMo improves mIoU by 1.9\% points while operating under 70.0\% lower GFLOPs, significantly reduced (70.6\%) latency, and 46.9\% less parameters.
Unlike purely convolutional designs, SiConMo explicitly integrates global context within a compact bottleneck, allowing more effective semantic reasoning without increasing model complexity.
Furthermore, under strict real-time constraints, SiConMo outperforms recent lightweight hybrid models including TopFormer and SeaFormer at comparable computational budgets.
Overall, these results demonstrate that SiConMo provides a strong balance between segmentation accuracy and efficiency for real-time semantic segmentation.
\textit{Details of datasets, evaluation protocols, and implementation settings are provided in the supplementary material.}%

\begin{figure}[t]
  \centering
  \includegraphics[width=\linewidth]{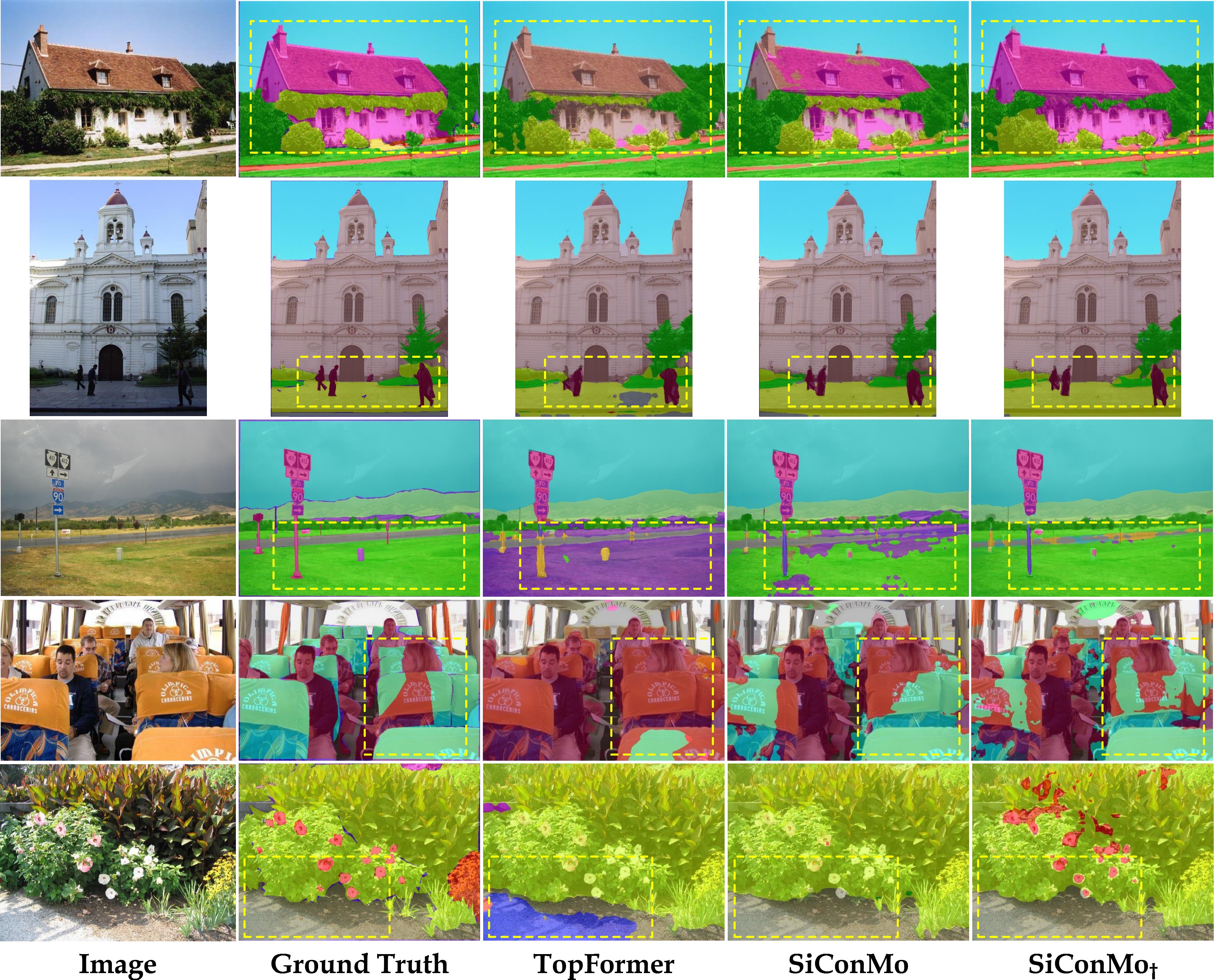}
  \vspace{-8mm}
  \caption{Visual results on the ADE20K validation set. The proposed SiConMo model produces high-quality segmentation maps with improved spatial consistency and fine-detail preservation, compared to TopFormer~\cite{zhang2022topformer}.}
  \label{fig:proposed_method_visual_results}
  \vspace{-5mm}
\end{figure}

\textbf{Visual Results:}
Qualitative comparisons on the ADE20K validation set are presented in \Cref{fig:proposed_method_visual_results}. Compared with representative lightweight baselines such as TopFormer~\cite{zhang2022topformer}, SiConMo produces segmentation maps with improved boundary delineation, finer structural details, and enhanced spatial consistency. In particular, small objects and complex scene layouts are better preserved, while over-smoothing artifacts are reduced. These visual observations are consistent with the quantitative improvements reported in \Cref{tab:ade20k_mega}(a).

\vspace{-3mm}
\subsubsection{Ablation Study}
\label{sec:ablation}
\noindent \textbf{Effects of BDC Branch Design:}
The upper part of \Cref{tab:ade20k_mega}(c) evaluates the contribution of individual components in the Branched Depthwise Convolution (BDC) branch.
Adding a $1\times1$ depthwise convolution to the $3\times3$ baseline provides a small accuracy gain, while introducing a parallel $3\times3$ depthwise separable branch yields a substantial improvement, highlighting the benefit of multi-scale feature modeling.
Channel attention further enhances performance with negligible computational overhead.

\noindent \textbf{Effect of Trans-BDC Block and Edge Guidance:}
The lower part of \Cref{tab:ade20k_mega}(c) analyzes the interaction between the lightweight ViT branch and the BDC components.
The ViT branch alone significantly improves segmentation accuracy, and progressively adding BDC components consistently boosts performance, confirming their complementarity.
Incorporating gradient magnitude and edge maps provides a modest additional gain, improving boundary localization.
As shown in \Cref{tab:ade20k_mega}(d), Canny-based edge guidance increases computation without notable accuracy benefits.
Overall, the ablation results validate the effectiveness of each component in the Trans-BDC design.
\vspace{-2mm}

\begin{table}[t]
\small
  \caption{Results on PASCAL Context test set~\cite{mottaghi2014role}.
  }
  \label{tab:results_pascal_context}
\setlength{\tabcolsep}{2.5pt}
\renewcommand{\arraystretch}{0.95}
  \centering
  \resizebox{\columnwidth}{!}{%
  \begin{tabular}{c|c|c|c|c}
    \hline
    Methods & Backbone & mIoU$^{59}$ & mIoU$^{60}$ & GFLOPs \\
    \hline
    DeepLabV3+~\cite{chen2018encoder} & ENet-s16~\cite{paszke2016enet} & 43.07 & 39.19 & 23.00 \\
    DeepLabV3+~\cite{chen2018encoder} & MobileNetV2-s16~\cite{sandler2018mobilenetv2} & 42.34 & 38.59 & 22.24 \\
    LR-ASPP~\cite{howard2019searching} & MobileNetV3-s16~\cite{howard2019searching} & 38.02 & 35.05 & 2.04 \\
    TopFormer~\cite{zhang2022topformer} & TopFormer-T~\cite{zhang2022topformer} & 40.39 & 36.41 & 0.53 \\
    SeaFormer \cite{wan2023seaformer} & SeaFormer-T~\cite{wan2023seaformer} & 41.49 & 37.27 & 0.51 \\
    \hline
    \rowcolor{gray!20}
    SiConMo (Ours) & Ours & 41.84 & 37.49 & 0.47 \\
    \rowcolor{gray!20}
    SiConMo$_\dagger$ (Ours) & Ours & 41.85 & 37.78 & 0.49 \\
      \hline
    \end{tabular}
    }
    \vspace{-4mm}
\end{table}

\begin{table}[t]
\small
  \caption{Results on Cityscapes validation set~\cite{cordts2016cityscapes}.
  }
  \label{tab:results_cityscapes}
\renewcommand{\arraystretch}{0.90}
  \centering
  \resizebox{\columnwidth}{!}{%
  \begin{tabular}{c|c|c|c}
    \hline
    Methods & Encoder & mIoU & GFLOPs \\
    \hline
    PSPNet~\cite{zhao2017pyramid} & MobileNetV2~\cite{sandler2018mobilenetv2} & 70.2 & 423.4 \\
    FCN~\cite{long2015fully} & MobileNetV2~\cite{sandler2018mobilenetv2} & 61.5 & 317.1 \\
    SegFormer~\cite{xie2021segformer} & MiT-B0~\cite{xie2021segformer} & 71.9 & 17.7 \\
    L-ASPP~\cite{chen2018encoder} & MobileNetV2~\cite{sandler2018mobilenetv2} & 72.7 & 12.6 \\
    LR-ASPP~\cite{howard2019searching} & MobileNetV3-Large~\cite{howard2019searching} & 72.4 & 9.7 \\
    LR-ASPP~\cite{howard2019searching} & MobileNetV3-Small~\cite{howard2019searching} & 68.4 & 2.9 \\
    TopFormer~\cite{zhang2022topformer} & TopFormer-T~\cite{zhang2022topformer} & 66.5 & 1.2 \\
    SeaFormer \cite{wan2023seaformer}&SeaFormer-T~\cite{wan2023seaformer} & 66.8 & 1.2 \\
    \hline
    \rowcolor{gray!20}
    SiConMo (Ours) & Ours & 68.0 & 1.2 \\
    \rowcolor{gray!20}
    SiConMo$_\dagger$ (Ours) & Ours & 68.2 & 1.2 \\
      \hline
    \end{tabular}
    }
    \vspace{-6mm}
\end{table}
\begin{table}[t]
\small
  \caption{Results on COCO-Stuff test set~\cite{caesar2018coco}.
  }
  \label{tab:results_coco_stuff}
  \centering
\renewcommand{\arraystretch}{0.90}
  \resizebox{\columnwidth}{!}{%
  \begin{tabular}{c|c|c|c}
    \hline
    Methods & Encoder & mIoU & GFLOPs \\
    \hline
    PSPNet~\cite{zhao2017pyramid} & MobileNetV2-s8~\cite{sandler2018mobilenetv2} & 30.14 & 52.94 \\
    DeepLabV3+~\cite{chen2018encoder} & EfficientNet-s16~\cite{tan2019efficientnet} & 31.45 & 27.10 \\
    DeepLabV3+~\cite{chen2018encoder} & MobileNetV2-s16~\cite{sandler2018mobilenetv2} & 29.88 & 25.90 \\
    LR-ASPP~\cite{howard2019searching} & MobileNetV3-s16~\cite{howard2019searching} & 25.16 & 2.37 \\
    TopFormer~\cite{zhang2022topformer} & TopFormer-T~\cite{zhang2022topformer} & 28.34 & 0.64 \\
    SeaFormer \cite{wan2023seaformer} & SeaFormer-T~\cite{wan2023seaformer} & 29.24 & 0.62  \\
    \hline
    \rowcolor{gray!20}
    SiConMo (Ours) & Ours & 29.24 & 0.58 \\
    \rowcolor{gray!20}
    SiConMo$_\dagger$ (Ours) & Ours & 29.26 & 0.60 \\
      \hline
    \end{tabular}
    }
    \vspace{-6mm}
\end{table}

\subsection{Results PASCAL Context, Cityscapes, COCO-Stuff}
\vspace{-1.5mm}
As summarized in \Cref{tab:results_pascal_context}, \Cref{tab:results_cityscapes}, and \Cref{tab:results_coco_stuff}, SiConMo demonstrates a consistent and favorable accuracy–efficiency trade-off across diverse semantic segmentation benchmarks. On the PASCAL Context dataset, SiConMo delivers strong mIoU results under both 59-class and 60-class evaluation settings, substantially outperforming efficient CNN-based baselines such as LR-ASPP while requiring 76.9\% less GFLOPs. In comparison with lightweight Transformer-based models, SiConMo consistently attains higher accuracy with equal or lower computational cost, highlighting the effectiveness of its bottleneck-aware design in modeling both local structure and global context. On Cityscapes, SiConMo achieves competitive segmentation accuracy under strict real-time constraints, outperforming recent lightweight hybrid models such as TopFormer and SeaFormer while operating at the same computational budget. Compared to higher-capacity CNN- and Transformer-based approaches, SiConMo reduces computational cost by over two orders of magnitude while maintaining comparable performance.
Similarly, on COCO-Stuff, SiConMo achieves accuracy comparable to high-capacity DeepLabV3+ variants while reducing computational overhead by more than 97\%. Under real-time settings, it matches or surpasses recent lightweight baselines, including TopFormer and SeaFormer, with lower GFLOPs. 
These results confirm that SiConMo generalizes across datasets with varying complexity and class diversity, offering a strong accuracy–efficiency trade-off.

\vspace{-3mm}
\subsection{Object Detection}
\vspace{-2mm}
\label{subsec:object_detection}
To further assess the generalization capability of the proposed method, we conducted object detection experiments on the COCO dataset. COCO consists of 118K training images, 5K validation images, and 20K test images. For these experiments, we adopted RetinaNet, a one-stage detector, using SiConMo$_\dagger$ as the backbone to construct a feature pyramid at multiple scales.
All models were trained on the train2017 split for 12 epochs with ImageNet-pretrained weights and evaluated on the val2017 set.
As summarized in \Cref{tab:ade20k_mega}(e) and detailed in \Cref{tab:ade20k_mega}(f), the proposed method demonstrates strong performance across standard detection metrics, indicating its robustness and effectiveness in downstream tasks. These results further suggest that the backbone learned via our approach generalizes well beyond semantic segmentation.
\vspace{-4mm}

\section{Conclusion}
\vspace{-2mm}
In this work, we showed that simplicity can be an effective alternative to architectural complexity for lightweight semantic segmentation. We introduced SiConMo, a lightweight and bottleneck-aware framework that integrates CNN and Transformer components through a streamlined design. By incorporating token pyramid extraction, hybrid local–global context modeling, and efficient feature merging, SiConMo balances segmentation accuracy and computational efficiency. Experiments on ADE20K, PASCAL Context, Cityscapes, and COCO-Stuff demonstrate that SiConMo achieves competitive or improved performance compared to more complex models while operating under strict efficiency constraints.

\bibliographystyle{IEEEbib}
\bibliography{strings,refs}

\onecolumn
\newpage
\twocolumn
\section*{\centering ------ Supplementary Material ------}

\begin{figure*}
\centering
\includegraphics[width=\linewidth]{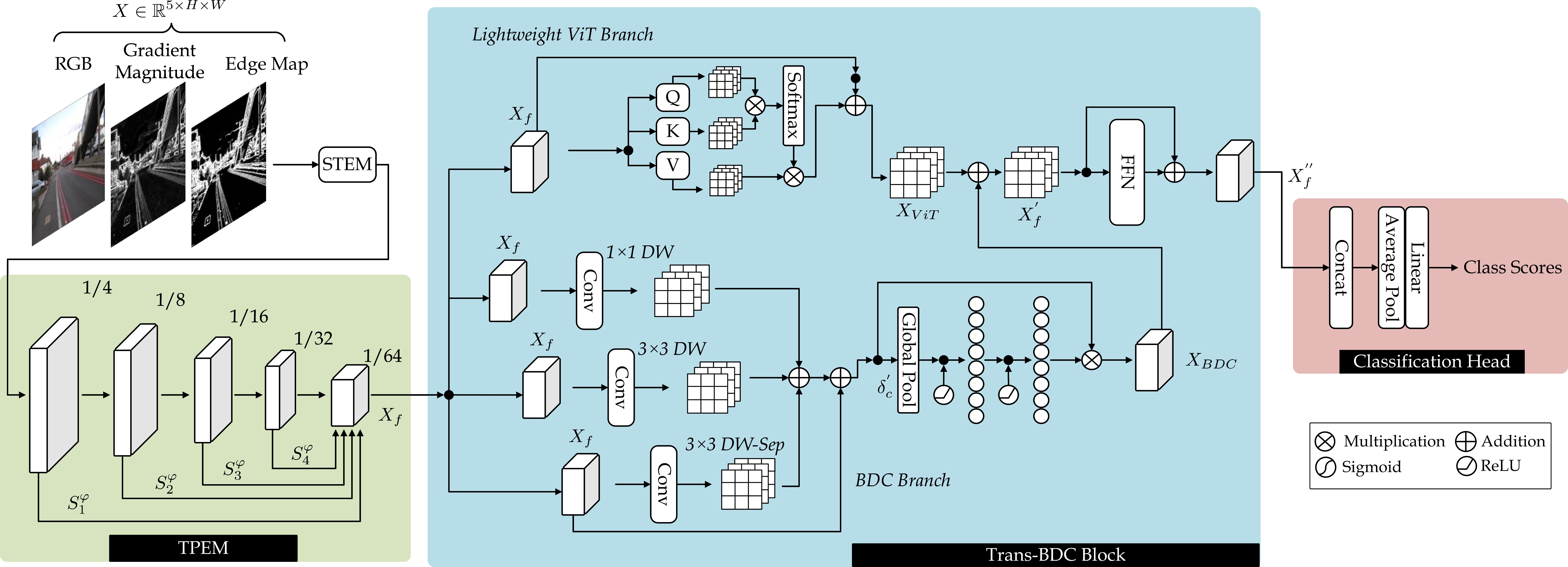}
\caption{The architecture of the proposed SiConMo model for the task of image classification.}
\label{fig:proposed_method_classification}
\end{figure*}

\section{Datasets and Measures}
We conduct the experiments on four benchmark datasets: ADE20K~\cite{zhou2017scene}, PASCAL Context~\cite{mottaghi2014role}, CityScapes~\cite{cordts2016cityscapes} and COCO-Stuff~\cite{caesar2018coco}.
In total, \textbf{ADE20K}~\cite{zhou2017scene} dataset consists of 25000 images, containing 150 class categories. The division of dataset is as follows: 20000 images for training, 2000 images for validation and 3000 images for testing.
The \textbf{PASCAL Context}~\cite{mottaghi2014role} dataset contains 1 background, and 59 semantic labels. Out of 10103 images in total, training and testing scene images are split into 4998 and 5105, respectively.
The \textbf{CityScapes}~\cite{cordts2016cityscapes} dataset consists of 19 fine class annotations. 2975 images are taken into account for training and 500 images for validation/testing.
Pixel-level stuff annotations were applied on COCO dataset for augmentation, resulting in \textbf{COCO-Stuff}~\cite{caesar2018coco} dataset. From COCO dataset, 10000 images are picked, where 9000 images are considered for training and 1000 images for testing.
Following the recent literature~\cite{cavagnero2024pem,zhang2022topformer} we report the results using the standard common measures: Mean Intersection over Union (mIoU) for segmentation accuracy, Giga Floating Point Operations per Second (GFLOPs), latency and number of parameters. 
\vspace{-2mm}
\section{Implementation Details}
Our implementation is based on PyTorch and MMSegmentation toolbox. All the models are first pretrained on ImageNet-1K dataset, including our proposed SiConMo model. Further, they are fine-tuned on semantic segmentation datasets. Batch-Normalization layers are used after almost each convolution layer except the last output layer. 
For the \textbf{ADE20K} dataset, we follow data augmentations same as in \cite{xie2021segformer} for fair comparison. Moreover, for ADE20K dataset, we use batch size 16, and 160K scheduler by following \cite{xie2021segformer} and \cite{zhang2022topformer}. Various augmentations have been applied such as random scaling, random cropping, random horizontal flip, random resize, etc. For the \textbf{CityScapes} dataset, the same data augmentations are followed as in \cite{xie2021segformer,zhang2022topformer}. Images are resized and rescaled with same crop size i.e. $1024 \times 512$. It should be noted that for all datasets and models, we set $1.2 \times 10^{-4}$ as the initial learning rate with  the weight decay set to $0.01$. However, the initial learning rate for the CityScapes dataset is set to $3 \times 10^{-4}$. 
For the \textbf{PASCAL Context} and \textbf{COCO-Stuff} datasets, 80K training iterations are implemented. Additionally, for both datasets, the same data augmentations and training settings are incorporated as in MMSegmentation and the training images are resized and cropped to $512 \times 512$ and $480 \times 480$ for COCO-Stuff and PASCAL Context datasets, respectively.

\section{ImageNet Pre-training} 
\label{sec:imagenet_pretraining}

To ensure a fair comparison, SiConMo is initialized with weights pre-trained on the ImageNet-1K dataset. As illustrated in \Cref{fig:proposed_method_classification}, the classification framework employs global average pooling followed by a linear projection to generate class scores, effectively utilizing high-level semantic representations from the backbone. For $224 \times 224$ inputs, the token resolution within the Trans-BDC block is configured to $\tfrac{1}{64} \times \tfrac{1}{64}$ of the image size. Quantitative results on ImageNet-1K are presented in \Cref{tab:results_imagenet1k}, demonstrating that SiConMo attains competitive accuracy while maintaining an extremely lightweight computational footprint.

\begin{table}[h]
  \caption{SiConMo results for ImageNet classification.
  }
  \label{tab:results_imagenet1k}
  \centering
  \resizebox{\columnwidth}{!}{%
  \begin{tabular}{c|c|c|c|c}
    \hline
    Method & Input Size & Top-1 Accuracy(\%) & GFLOPs & Parameters \\
    \hline
    \rowcolor{gray!20}
    SiConMo (Ours) & $224\times224$ & 66.2 & 0.13 & 1.79M \\
    \rowcolor{gray!20}
    SiConMo$_\dagger$ (Ours) & $224\times224$ & 66.4 & 0.13  & 1.79M  \\
      \hline
    \end{tabular}
    }
\end{table}

\section{Detailed Network Structure} 
\label{sec:network_details}

This section presents the detailed architecture of the proposed SiConMo model, designed to achieve real-time semantic segmentation while maintaining a lightweight and bottleneck-aware structure. The full architecture is summarized in~\Cref{tab:architectural_details_SiConMo}. SiConMo strategically introduces and integrates simple yet effective modules to balance efficiency and representational power: the Token Pyramid Extraction Module (TPEM) captures multi-scale local features, the Trans-BDC block models long-range dependencies without excessive complexity, and the Feature Merging Module (FMM) adaptively combines spatial and contextual information. 

In the table, $dw$ and $dw.sep$ indicate depth-wise and depth-wise separable convolutions, respectively, while N, H, and T denote the number of blocks, attention heads, and target channels. 
The table provides the architectural details of the network, based on an input resolution of $512 \times 512$, illustrating the layer properties and output resolutions at this size.
This modular and bottleneck-conscious design highlights how simplicity, when carefully applied, can achieve robust feature extraction and context modeling.

\begin{table}
  \caption{Architectural details of SiConMo model. For input resolution of $512 \times 512$.
  }
  \label{tab:architectural_details_SiConMo}
  \centering
  \resizebox{\columnwidth}{!}{%
  \begin{tabular}{@{}c|c|c|c|c|c|c@{}}
    \hline
    Stage & \multicolumn{5}{c|}{Details} & Output Resolution \\
    \cline{2-6}
    \cline{2-5}
 & layer & kernel size & expand ratio & output channels & stride \\
 \hline
 Stem & Conv & 3 & - & 16 & 2 & \\
   & MobileNetV2 & 3 & 1 & 16 & 1 & $256 \times 256$\\
 \hline
 & MobileNetV2 & 3 & 4 & 16 & 2 \\
 & MobileNetV2 & 3 & 3 & 16 & 1 & $128 \times 128$ \\
 \cline{2-7}
 & MobileNetV2 & 5 & 3 & 32 & 2 \\
 TPEM & MobileNetV2 & 5 & 3 & 32 & 1 & $64 \times 64$\\
 \cline{2-7}
 & MobileNetV2 & 3 & 3 & 64 & 2 \\
 & MobileNetV2 & 3 & 3 & 64 & 1 & $32 \times 32$\\
 \cline{2-7}
 & MobileNetV2 & 5 & 6 & 96 & 2 \\
 & MobileNetV2 & 5 & 6 & 96 & 1 & $16 \times 16$\\
 \hline
 \hline
  & \multicolumn{5}{c|}{Trans-BDC Block} & \\
 \hline
 BDC Branch & $3\times3_{dw}$ & 3 & - & 208 & 1 &  \\
  & $1\times1_{dw}$ & 1 & - & 208 & 1 &  \\
  & $3\times3_{dw.sep}$ & 3 , 1 & - & 208 & 1 &  \\
  \cline{2-6}
   & \multicolumn{5}{c|}{N=4} & $8 \times 8$ \\
  \hline
  ViT Branch & \multicolumn{5}{c|}{N=4, H=4} & $8 \times 8$ \\
  \hline
  \hline
  FMM & \multicolumn{5}{c|}{T=160} & 8$^2$, 16$^2$, 32$^2$, 64$^2$ \\
  \hline
  GFLOPs & \multicolumn{5}{c|}{-} & 0.6 \\
  \hline
    \end{tabular}
    }
\end{table}

\section{Details of Feed-Forward Network} 
\label{sec:fnn_details}

\begin{figure}[h]
  \centering
  \includegraphics[width=3.54cm]{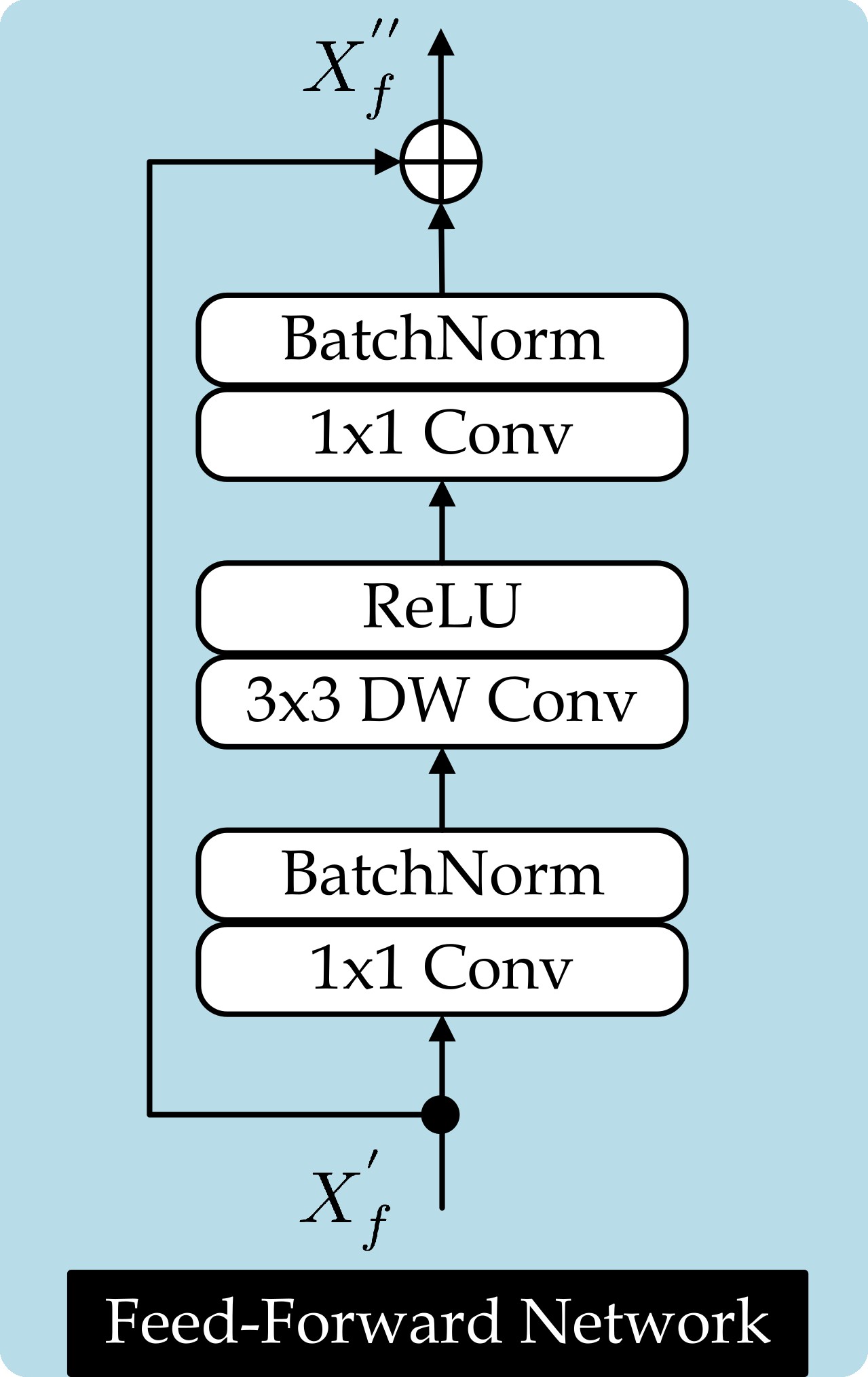}
  \caption{Design of Feed-Forward Network.
  }
  \label{fig:feed_forward_network}
\end{figure}

In SiConMo, the unified features from the BDC branch and Lightweight ViT branch are processed through a lightweight Feed-Forward Network (FFN) to enhance the representation while preserving efficiency. The FFN integrates a depth-wise convolution between two $1 \times 1$ convolution layers, enabling effective local feature refinement and capturing contextual information without incurring substantial computational overhead. An expansion factor of two is used to balance the expressiveness of the features with the lightweight design, complementing the bottleneck-aware Trans-BDC block. This carefully designed FFN contributes to SiConMo’s ability to model global and local semantics efficiently, reinforcing the overall simplicity and robustness of the network. The architecture of the Feed-Forward Network is illustrated in~\Cref{fig:feed_forward_network}.

\begin{figure*}[t]
  \centering
  \includegraphics[width=10.5cm]{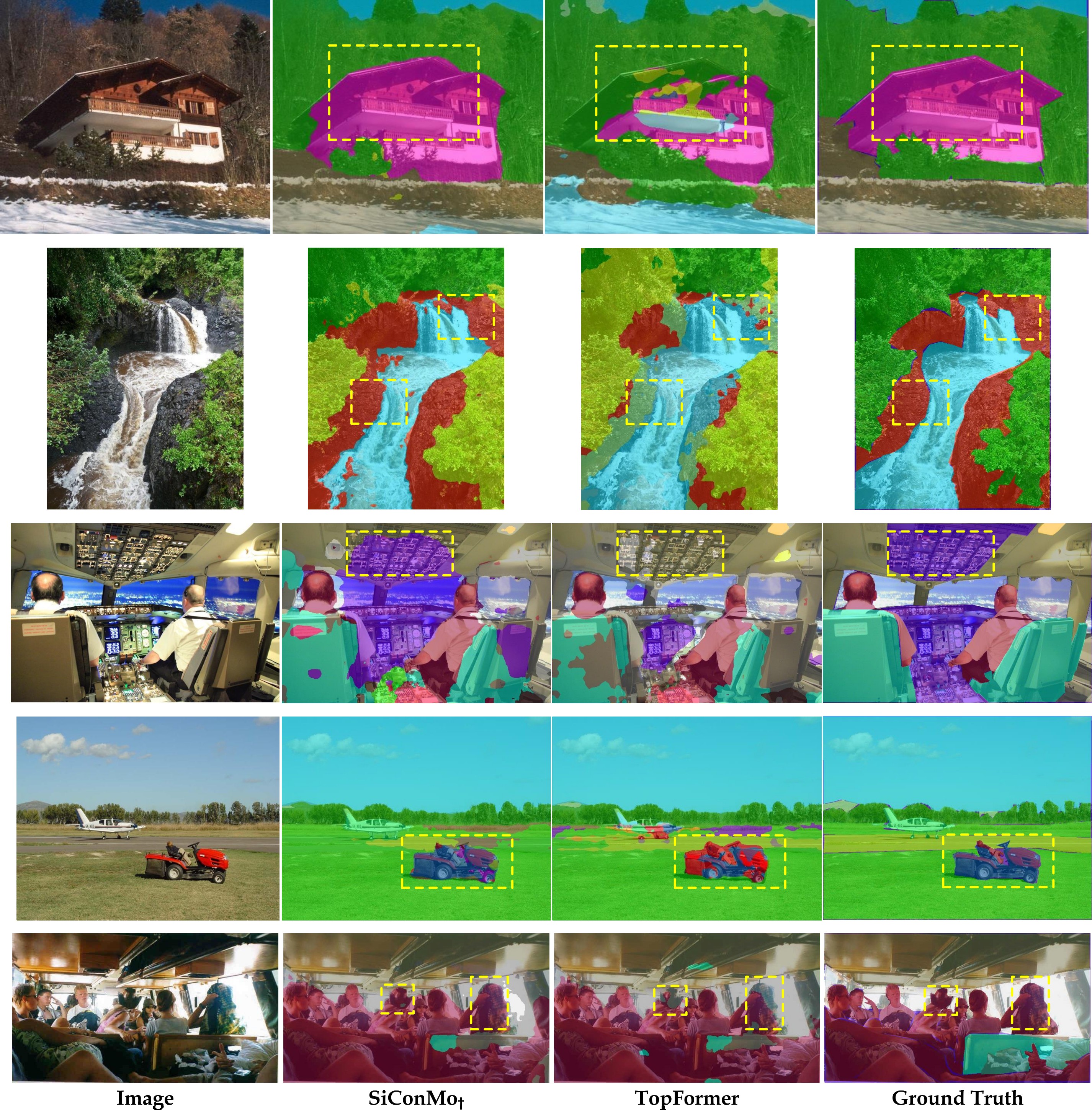}
  \caption{Visual results on the ADE20K validation set. Our proposed model illustrates superior consistency with the ground truth segmentation results, highlighting its robustness. 
  }
  \label{fig:proposed_method_visual_results_supp}
\end{figure*}

\section{Gradient Magnitude and Edge Map (GME) Generation}
For the enhanced SiConMo$_\dagger$ variant, we augment the RGB input with Gradient Magnitude and Edge Maps (GME) to improve structural awareness.

Given an RGB input image $X \in \mathbb{R}^{H \times W \times 3}$, we first convert it to grayscale using the standard luminance formulation:
\begin{equation}
X_{gray} = 0.2989R + 0.5870G + 0.1140B.
\end{equation}

We then compute horizontal and vertical gradients using Sobel operators:
\begin{equation}
S_x =
\begin{bmatrix}
-1 & 0 & 1\\
-2 & 0 & 2\\
-1 & 0 & 1
\end{bmatrix},
\qquad
S_y =
\begin{bmatrix}
-1 & -2 & -1\\
0 & 0 & 0\\
1 & 2 & 1
\end{bmatrix}.
\end{equation}

The horizontal and vertical gradients are obtained via convolution:
\begin{equation}
G_x = X_{gray} * S_x,
\qquad
G_y = X_{gray} * S_y,
\end{equation}
where $*$ denotes convolution.

The gradient magnitude map is computed as:
\begin{equation}
G_m = \sqrt{G_x^2 + G_y^2}.
\end{equation}

An edge map is further generated using mean-based thresholding:
\begin{equation}
E =
\begin{cases}
1, & G_m > \mu(G_m) \\
0, & \text{otherwise},
\end{cases}
\end{equation}
where $\mu(G_m)$ denotes the mean gradient magnitude.

Finally, the gradient magnitude and edge maps are concatenated with the RGB image to form a five-channel input representation:
\begin{equation}
X' = [R, G, B, G_m, E].
\end{equation}

No additional normalization is applied to the generated GME channels beyond the standard RGB preprocessing. For SiConMo$_\dagger$, reported latency includes the computational cost of GME generation, since Sobel gradient magnitude and edge maps are computed within the model forward pass during inference.

\section{Visual Results} 
\label{sec:visual_results_supplementary}

~\Cref{fig:proposed_method_visual_results_supp} presents additional qualitative results of the proposed SiConMo model on the ADE20K validation set. For comparison, we show the original images, ground-truth annotations, predictions from TopFormer, and outputs from SiConMo$_\dagger$. The visualizations highlight that SiConMo produces more consistent and accurate segmentation, effectively capturing both fine-grained details and global context. These results underscore the robustness of our lightweight, bottleneck-aware design and demonstrate that careful architectural simplicity can outperform more complex models in real-world segmentation scenarios.

\section{Limitations and Future Work} 
\label{sec:limitations_futurework}
While the proposed SiConMo model demonstrates strong performance and efficiency for lightweight semantic segmentation, several limitations remain. A primary constraint, shared with many lightweight architectures, is its reliance on ImageNet-1K pre-training; omitting this step results in a noticeable drop in segmentation accuracy. Additionally, although the model is designed for efficiency, extreme edge cases with highly complex scenes or unusual object distributions may still challenge its predictive capabilities.

Future work will focus on enhancing the robustness and adaptability of SiConMo across diverse datasets and deployment scenarios, without compromising its lightweight design. We also aim to explore techniques for further reducing pre-training dependency, improving generalization to unseen domains, and extending the model for practical real-world applications, where balancing simplicity, accuracy, and efficiency remains crucial. These directions aim to position SiConMo as a flexible and practical framework for both research and applied vision tasks.


\end{document}